\documentclass{article}
\usepackage[preprint]{neurips_2025}
\usepackage{graphicx}
\usepackage{tikz}
\usetikzlibrary{shapes,arrows,positioning,calc,decorations.pathmorphing,backgrounds,fit,patterns}
\usepackage{amsmath}
\usepackage{amssymb}
\usepackage{amsthm}
\usepackage{natbib}
\usepackage{hyperref}
\usepackage{booktabs}
\usepackage{algorithm}
\usepackage{algorithmic}
\usepackage[dvipsnames]{xcolor}
\definecolor{gold}{RGB}{218, 165, 32}
\usepackage{adjustbox}

\newtheorem{definition}{Definition}
\newtheorem{example}{Example}

\title{{Why are there many equally good models?} \\\textit{An Anatomy of the Rashomon Effect}
 }
\author{Harsh Parikh\thanks{Contact: \texttt{harsh.parikh@yale.edu}\\I want to thank Prof Cynthia Rudin and Srikar Katta for discussions and inputs.\\ This is a working manuscript.}\\
Yale University}
\date{}

\begin{document}

\maketitle

\begin{abstract}
The Rashomon effect---the existence of multiple, distinct models that achieve nearly equivalent predictive performance---has emerged as a fundamental phenomenon in modern machine learning and statistics. In this paper, we explore the causes underlying the Rashomon effect, organizing them into three categories: \textit{statistical} sources arising from finite samples and noise in the data-generating process; \textit{structural} sources arising from non-convexity of optimization objectives and unobserved variables that create fundamental non-identifiability; and \textit{procedural} sources arising from limitations of optimization algorithms and deliberate restrictions to suboptimal model classes. We synthesize insights from machine learning, statistics, and optimization literature to provide a unified framework for understanding why the multiplicity of good models arises. A key distinction emerges: statistical multiplicity diminishes with more data, structural multiplicity persists asymptotically and cannot be resolved without different data or additional assumptions, and procedural multiplicity reflects choices made by practitioners. Beyond characterizing causes, we discuss both the challenges and opportunities presented by the Rashomon effect, including implications for inference, interpretability, fairness, and decision-making under uncertainty. 
\end{abstract}

\section{Introduction}
\label{sec:introduction}

Practitioners have long observed that when fitting models to data, many substantially different models often achieve nearly identical performance. A random forest, a neural network, and a logistic regression fitted to the same dataset may all yield comparable predictive accuracy. More intriguingly, these models may rely on entirely different variables, suggest different causal mechanisms, and produce different predictions for individual cases---all while being indistinguishable by standard performance metrics \citep{fisher2019all, marx2020predictive}. This phenomenon---the existence of multiple, distinct models that are equally good---has come to be known as the \emph{Rashomon effect}. The term was coined by \citet{breiman2001statistical}, borrowing from Akira Kurosawa's 1950 film \emph{Rashomon}, in which four witnesses provide contradictory yet equally plausible accounts of the same crime. Akin to the film, the Rashomon effect in statistical modeling challenges the assumption that data uniquely determine a best model.

The Rashomon effect is not merely a theoretical curiosity. In domains such as criminal justice, healthcare, and lending, algorithmic decisions affect individual lives. When multiple models are equally justified by the data yet produce different outcomes for specific individuals, the choice among them becomes arbitrary in a troubling sense \citep{black2022model}. A defendant's bail decision, a patient's treatment recommendation, or an applicant's loan approval may depend not on the evidence but on which equally-good model happened to be selected. Beyond individual decisions, the Rashomon effect complicates scientific inference: when different models highlight different variables as important, data alone cannot tell us which variables truly matter for the outcome \citep{watson2023predictive, donnelly2023rashomon}. Different research teams analyzing the same dataset may reach different conclusions simply because they selected different models from the Rashomon set, contributing to the reproducibility challenges that have troubled many scientific fields \citep{ioannidis2005most}.

Yet the Rashomon effect is not purely a source of consternation. Recent work has revealed that model multiplicity can be a resource rather than merely a liability. \citet{rudin2024amazing} argue that the Rashomon effect ``unlocks a treasure trove of information about the relationship of real datasets to families of predictive models.'' When many good models exist, some are likely to be simple, interpretable, fair, or otherwise desirable along dimensions beyond predictive accuracy \citep{semenova2022existence}. The existence of large Rashomon sets often implies that there is no accuracy-interpretability tradeoff for a given problem: an interpretable model exists that performs as well as any black-box alternative \citep{rudin2019stop}. Furthermore, characterizing the full Rashomon set rather than selecting a single model enables more honest uncertainty quantification, more robust variable importance estimates, and more principled approaches to incorporating domain knowledge \citep{fisher2019all, coker2021theory}.

To harness the opportunities presented by the Rashomon effect while mitigating its risks, we must first understand why it occurs. What features of data, models, and learning procedures give rise to model multiplicity? Under what conditions should we expect Rashomon sets to be large versus small? How do different causes interact, and what do they imply for practice? These questions motivate the present paper.

\paragraph{Contributions and Overview.}
This paper provides a systematic examination of the causes underlying the Rashomon effect, synthesizing insights from machine learning, statistics, econometrics, and causal inference. We organize the causes into three categories that differ in their nature and implications for practice.

\textbf{Statistical sources} (Section~\ref{sec:finite-samples-noise}) arise from the finiteness of samples and noise in the data-generating process. With finite data, many models may be statistically indistinguishable even when a unique population-optimal model exists. \citet{semenova2023path} establish that noise in outcomes creates a causal pathway to large Rashomon sets: higher noise leads to larger generalization gaps, which force practitioners toward simpler model classes, which in turn exhibit larger Rashomon ratios. Crucially, statistical multiplicity diminishes as sample size grows---in the limit of infinite data, this source of multiplicity vanishes.

\textbf{Structural sources} (Sections~\ref{sec:nonconvex}--\ref{sec:missing}) arise from properties of the optimization landscape and fundamental limitations on what can be learned from observable data. Non-convexity of optimization objectives (Section~\ref{sec:nonconvex}) may create multiple minima, saddle points, and plateau regions that persist regardless of sample size. We highlight that non-convex objectives arise not only in deep learning but also in causal inference---for example, the variance minimization objective studied by \citet{parikh2024missing} for characterizing underrepresented populations exhibits complex non-convexity that precludes standard optimization approaches. Unobserved data, including missing covariates, latent variables, and unmeasured confounders (Section~\ref{sec:missing}), creates multiplicity through non-identifiability. A key contribution of our paper is connecting the Rashomon effect to the \emph{partial identification} framework developed in econometrics \citep{manski2003partial, tamer2010partial}. We argue that partial identification is a special case of the Rashomon effect where the tolerance $\epsilon$ equals zero and multiplicity persists regardless of sample size. Structural multiplicity cannot be resolved by collecting more data of the same type---it requires either different data (e.g., experiments, new measurements) or additional assumptions.

\textbf{Procedural sources} (Sections~\ref{sec:optimizer}--\ref{sec:suboptimal}) arise from choices made in the modeling pipeline rather than from the data or problem structure. Limitations of optimization algorithms (Section~\ref{sec:optimizer})---including initialization dependence, stochastic gradient noise, early stopping, and computational constraints---cause different runs to discover different members of the Rashomon set even when a unique optimum exists in principle. This source is often undervalued but plays a critical role in practice, as documented by \citet{damour2022underspecification}. Deliberate restriction to suboptimal model classes (Section~\ref{sec:suboptimal})---through regularization, architectural constraints, and interpretability requirements---can create multiplicity when the true optimum is excluded from consideration. Procedural multiplicity could in principle be reduced through better algorithms or broader model classes, though practical constraints often make this infeasible.

We conclude with a discussion of the interactions among causes, implications for machine learning practice, and open questions for future research (Section~\ref{sec:discussion}). Throughout, we emphasize that the Rashomon effect is neither purely good nor purely bad---it is a fundamental feature of learning from data that carries both risks and opportunities depending on context.

\section{Formal Framework and Quantification}
\label{sec:framework}

This section establishes the formal definitions used throughout the paper and reviews methods for quantifying the Rashomon effect. We then provide illustrative examples demonstrating the phenomenon's ubiquity across modeling contexts.

\subsection{Definitions}

\begin{definition}[Rashomon Set]
\label{def:rashomon_set}
Let $\mathcal{F}$ denote a hypothesis class of models, $\mathcal{D} = \{(x_i, y_i)\}_{i=1}^n$ a dataset, and $L: \mathcal{F} \times \mathcal{D} \rightarrow \mathbb{R}^+$ a loss function. Let $f^* = \arg\min_{f \in \mathcal{F}} L(f, \mathcal{D})$ denote the empirical risk minimizer with optimal loss $L^* = L(f^*, \mathcal{D})$. The \textbf{Rashomon set} $\mathcal{R}(\epsilon, \mathcal{F}, \mathcal{D})$ for a given tolerance $\epsilon > 0$ is defined as:
\begin{equation}
\mathcal{R}(\epsilon, \mathcal{F}, \mathcal{D}) = \{f \in \mathcal{F} : L(f, \mathcal{D}) \leq L^* + \epsilon\}
\end{equation}
\end{definition}

The Rashomon set contains all models whose empirical risk falls within $\epsilon$ of the optimal achievable risk. When this set is large and contains structurally or functionally diverse models, we say that the learning problem exhibits a strong Rashomon effect \citep{fisher2019all, semenova2022existence}. The tolerance $\epsilon$ determines the strictness of the equivalence criterion.

\begin{definition}[Rashomon Ratio]
\label{def:rashomon_ratio}
The \textbf{Rashomon ratio} $\rho(\epsilon, \mathcal{F}, \mathcal{D})$ quantifies the relative size of the Rashomon set:
\begin{equation}
\rho(\epsilon, \mathcal{F}, \mathcal{D}) = \frac{|\mathcal{R}(\epsilon, \mathcal{F}, \mathcal{D})|}{|\mathcal{F}|}
\end{equation}
where $|\cdot|$ denotes an appropriate measure on the hypothesis class (e.g., cardinality for finite classes, or volume for continuous parameterizations).
\end{definition}

\citet{semenova2022existence} found that large Rashomon ratios correlate with problem tractability: when many models achieve near-optimal training performance, some typically generalize well.

\subsection{Illustrative Examples}

The Rashomon effect manifests across diverse modeling contexts.

\begin{example}[Linear Regression with Collinear Predictors]
Consider predicting a response $y$ using predictors $x_1$ and $x_2$ with $\text{cor}(x_1, x_2) \approx 0.99$. Models using primarily $x_1$, primarily $x_2$, or various linear combinations may achieve nearly identical MSE while implying fundamentally different predictor-response relationships \citep{hastie2009elements}. Small perturbations to the data can dramatically change coefficient estimates with minimal impact on predictive performance.
\end{example}

\begin{example}[Decision Trees]
\citet{xin2022exploring} developed TreeFARMS to enumerate sparse decision trees achieving near-optimal accuracy. For the COMPAS recidivism dataset, over 1,000 trees achieve performance within 1\% of optimal, each with substantially different structures and variable usage patterns. This multiplicity implies that variable importance rankings from any single tree may be misleading.
\end{example}

\begin{example}[Neural Networks]
\citet{damour2022underspecification} demonstrated that neural networks achieving similar held-out performance can exhibit dramatically different behavior under distribution shift. Even the choice of random seed substantially impacts model behavior, raising concerns about the arbitrary nature of deployment decisions.
\end{example}

\begin{example}[Cross-Model-Class Multiplicity]
\label{ex:fico}
On the FICO credit scoring dataset, \citet{rudin2024amazing} show that random forests (AUC 0.757), neural networks (0.792), and logistic regression (0.801) achieve comparable performance yet exhibit dramatically different variable reliance. Boosted trees rely heavily on 
\texttt{ExternalRisk} (18\% loss increase when permuted). Alternatively, logistic regression relies on \texttt{NetFractionRevolvingBurden} (23\% loss increase). Remarkably, simple interpretable models match black-box performance, illustrating that multiplicity spans model families.
\end{example}

\subsection{Metrics for Quantifying Multiplicity}
\label{subsec:metrics}

Several complementary metrics characterize different aspects of model multiplicity.

\paragraph{Size-Based Metrics.}
The Rashomon ratio (Definition~\ref{def:rashomon_ratio}) measures the fraction of the hypothesis class achieving near-optimal performance. For specific model classes, algorithms like TreeFARMS \citep{xin2022exploring} and methods for sparse GAMs \citep{zhong2023exploring} enumerate the Rashomon set exactly.

\paragraph{Prediction-Focused Metrics.}
\citet{marx2020predictive} introduced \textit{ambiguity}---the fraction of instances for which two models in the Rashomon set make conflicting predictions---and \textit{discrepancy}---the maximum disagreement between any model in the set and a reference model. \citet{semenova2022existence} introduced the \textit{pattern Rashomon ratio}, counting unique prediction patterns rather than models, to measure functional diversity. For continuous outputs, prediction variance $\text{Var}_{f \in \mathcal{R}}[f(x)]$ quantifies uncertainty attributable to model choice.

\paragraph{Explanation-Focused Metrics.}
\citet{fisher2019all} proposes \textit{model class reliance}, a metric for feature importance across models in the Rashomon set, as a range $[\text{MCR}^-, \text{MCR}^+]$. 
Variables with narrow intervals are robustly important; those with wide intervals have importance sensitive to model choice. \citet{donnelly2023rashomon} proposed the \textit{Rashomon Importance Distribution (RID)}, which marginalizes over both model multiplicity and sampling variability to produce stable importance estimates. Recent work by \citet{donnelly2025universe} extends these approaches to settings with unobserved confounding, providing the first variable importance bounds that simultaneously account for model multiplicity, finite-sample uncertainty, and omitted variables.

\paragraph{Information-Theoretic Metrics.}

\citet{hsu2022rashomon} introduced \textit{Rashomon capacity} to capture the volume and diversity of predictions within the Rashomon set to quantify ``predictive multiplicity'', quantifying in bits the uncertainty about predictions remaining after constraining to the Rashomon set.

The choice of metric depends on the analysis goal: size-based metrics assess whether multiplicity exists, prediction-focused metrics characterize impacts on individual decisions, and explanation-focused metrics assess robustness of scientific conclusions.

\subsection{Consequences for Inference and Practice}
\label{subsec:consequences}

The Rashomon effect has immediate consequences for inference and decision-making. Different models in the Rashomon set can suggest fundamentally different explanations: on the FICO dataset, different model classes rank variables in entirely different orders of importance, implying different causal narratives and policy interventions \citep{rudin2024amazing}. Without additional constraints, data alone cannot adjudicate among these explanations \citep{fisher2019all}.

Even when models agree on aggregate metrics, they may produce different predictions for individuals. \citet{marx2020predictive} found that more than 10\% of individuals receive different classifications depending on which model is selected---troubling for high-stakes decisions in criminal justice or healthcare \citep{black2022model}. \citet{cooper2024arbitrariness} showed that apparent fairness improvements can be artifacts of model selection.

The Rashomon effect connects to several related phenomena. Sensitivity to random seeds in deep learning \citep{damour2022underspecification} manifests the Rashomon effect at the optimization level. In econometrics, \textit{partial identification} \citep{manski2003partial, tamer2010partial} refers to situations where data constrain parameters to a set rather than a point. We argue below that partial identification is a special case of the Rashomon effect where multiplicity persists asymptotically.

These observations motivate our systematic examination of the \textit{causes} of the Rashomon effect: understanding why multiplicity arises determines how to respond.

\section{Finite Samples and Noisy Processes}
\label{sec:finite-samples-noise}

\textit{This section addresses statistical sources of multiplicity---those arising from the inherent uncertainty of learning from finite, noisy data. Unlike the structural and procedural sources discussed in subsequent sections, statistical multiplicity diminishes as sample size grows and vanishes in the limit of infinite data.}

\begin{figure}[htbp]
  \centering
  \begin{adjustbox}{max width=\textwidth}
    \begin{tikzpicture}[
        every node/.style={font=\small},
        arrow/.style={->, thick, >=stealth}
    ]
    
    \definecolor{statblue}{RGB}{70, 130, 180}
    \definecolor{statlightblue}{RGB}{200, 220, 240}
    
    \node[rectangle, rounded corners, draw=statblue, very thick, fill=statblue, text=white, 
          font=\bfseries\large, minimum width=4.5cm, minimum height=0.9cm] 
        (cause) at (5.5, 6.2) {Finite Samples \& Noise};
    
    \node[rectangle, rounded corners, draw=statblue, thick, fill=statlightblue!20, 
          minimum width=5.2cm, minimum height=5cm] (ill1) at (2.8, 2.8) {};
    \node[font=\bfseries, text=statblue] at (2.8, 4.9) {Different Models Fit Well};
    
    \begin{scope}[shift={(0.7, 1.0)}]
        \draw[->, thick, statblue!70] (0,0) -- (4,0) node[right, font=\footnotesize] {$x$};
        \draw[->, thick, statblue!70] (0,0) -- (0,3) node[above, font=\footnotesize] {$y$};
        
        \foreach \x/\y in {0.25/0.7, 0.45/0.95, 0.65/0.85, 0.85/1.15, 1.05/1.35, 
                           1.25/1.25, 1.45/1.55, 1.65/1.45, 1.85/1.7, 2.05/1.85, 
                           2.25/1.75, 2.45/2.0, 2.65/1.9, 2.85/2.15, 3.05/2.05, 
                           3.25/2.3, 3.45/2.2, 3.65/2.45} {
            \fill[statblue!70] (\x, \y) circle (2.5pt);
        }
        
        \draw[red!70!black, thick] (0.1, 0.65) -- (3.8, 2.5);
        
        \draw[green!50!black, thick, smooth] plot coordinates {
            (0.1, 0.7) (0.5, 0.85) (1.0, 1.1) (1.5, 1.4) (2.0, 1.7) 
            (2.5, 1.95) (3.0, 2.15) (3.5, 2.3) (3.8, 2.38)
        };
        
        \draw[orange!80!black, thick, smooth] plot coordinates {
            (0.1, 0.4) (0.3, 0.7) (0.6, 1.0) (1.0, 1.3) (1.5, 1.55) 
            (2.0, 1.75) (2.5, 1.92) (3.0, 2.08) (3.5, 2.22) (3.8, 2.32)
        };
        
        \node[font=\tiny, red!70!black] at (3.3, 2.7) {linear};
        \node[font=\tiny, green!50!black] at (2.4, 2.55) {quadratic};
        \node[font=\tiny, orange!80!black] at (1.5, 2.6) {log};
    \end{scope}
    
    
    \node[rectangle, rounded corners, draw=statblue, thick, fill=statlightblue!20, 
          minimum width=5.2cm, minimum height=5cm] (ill2) at (8.2, 2.8) {};
    \node[font=\bfseries, text=statblue] at (8.2, 4.9) {Indistinguishability Region};
    
    \begin{scope}[shift={(6.1, 1.0)}]
        \draw[->, thick, statblue!70] (0,0) -- (4,0) node[right, font=\footnotesize] {$\theta$};
        \draw[->, thick, statblue!70] (0,0) -- (0,3) node[above, font=\footnotesize] {$L(\theta)$};
        
        \draw[statblue!60, thick, dashed, smooth] plot coordinates {
            (0.2, 2.7) (0.6, 1.8) (1.0, 1.1) (1.4, 0.6) (1.8, 0.3) 
            (2.2, 0.25) (2.6, 0.45) (3.0, 0.85) (3.4, 1.45) (3.8, 2.2)
        };
        \node[font=\tiny, statblue!70] at (0.9, 2.5) {$L_{\text{pop}}(\theta)$};
        
        \fill[statblue] (2.1, 0.22) circle (3pt);
        \node[font=\tiny, statblue] at (2.1, -0.15) {$\theta^*$};
        
        \fill[red!20, opacity=0.6] 
            (1.0, 0) -- (1.0, 1.1) 
            to[out=0, in=180] (2.1, 0.25) 
            to[out=0, in=180] (3.2, 1.3) 
            -- (3.2, 0) -- cycle;
        \draw[red!50!black, thick, densely dotted] (1.0, 0) -- (1.0, 1.1);
        \draw[red!50!black, thick, densely dotted] (3.2, 0) -- (3.2, 1.3);
        
        \fill[green!30, opacity=0.7] 
            (1.65, 0) -- (1.65, 0.52) 
            to[out=0, in=180] (2.1, 0.25) 
            to[out=0, in=180] (2.55, 0.52) 
            -- (2.55, 0) -- cycle;
        \draw[green!50!black, thick] (1.65, 0) -- (1.65, 0.52);
        \draw[green!50!black, thick] (2.55, 0) -- (2.55, 0.52);
        
        \draw[<->, red!50!black, thick] (1.0, 1.3) -- (3.2, 1.3);
        \node[font=\tiny, red!50!black] at (2.1, 1.5) {small $n$};
        
        \draw[<->, green!50!black, thick] (1.65, 0.6) -- (2.55, 0.6);
        \node[font=\tiny, green!50!black] at (2.1, 0.8) {large $n$};
        
        \draw[gray, dashed] (0, 1.0) -- (3.9, 1.0);
        \node[font=\tiny, gray] at (3.6, 1.2) {$L^* + \epsilon$};
    \end{scope}

    \draw[arrow, statblue] (cause.south) -- ++(0,-0.25) -| (ill1.north);
    \draw[arrow, statblue] (cause.south) -- ++(0,-0.25) -| (ill2.north);
    
    \end{tikzpicture}
  \end{adjustbox}
  \caption{\textbf{Statistical sources} of the Rashomon effect arise from finite samples and noise. Left: With noisy data, fundamentally different models (linear, quadratic, logarithmic) may fit equally well and be statistically indistinguishable. Right: Even when the true loss $L_{\text{pop}}(\theta)$ has a unique minimum $\theta^*$, finite samples create an indistinguishability region where many models achieve similar empirical performance. This region shrinks as sample size increases.}
  \label{fig:rashomon-statistical}
\end{figure}
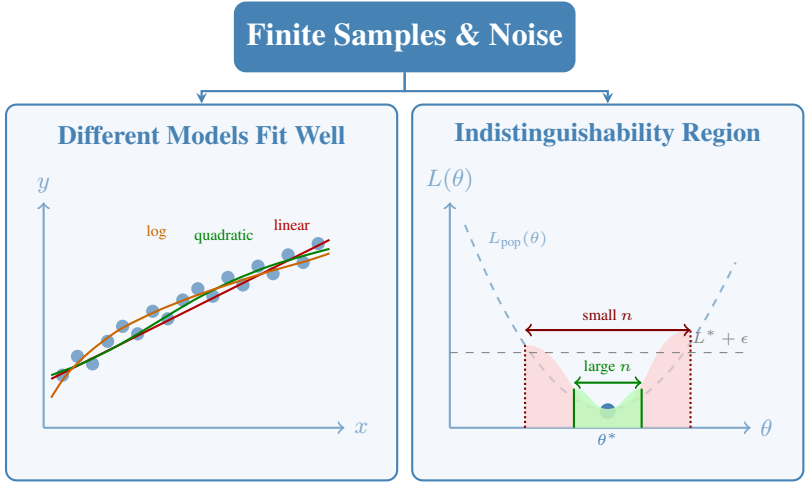

Even when a unique optimal model exists in the population, finite samples provide only imperfect information about the data-generating process. Many models may be statistically indistinguishable given available data, and inherent noise in outcomes limits how precisely any model can capture the true input-output relationship. These factors interact to determine both the size of the Rashomon set and the complexity of models that can generalize well.

\paragraph{Finite Sample Indistinguishability.}
With limited data, models with different population risks may achieve similar empirical risks. Standard uniform convergence results formalize this: for a model class $\mathcal{F}$ with VC dimension $d$, the gap between empirical and population risk is bounded by $O(\sqrt{d/n})$ with high probability \citep{hastie2009elements}. Models whose population risks differ by less than this bound are indistinguishable from finite samples. The Rashomon set thus includes not only models close to the empirical risk minimizer but also models close to the population risk minimizer---which may differ substantially when $n$ is small relative to model complexity.

\paragraph{Sensitivity to Data Selection.}
The particular sample observed profoundly shapes which models are learned. \citet{bouthillier2021accounting} systematically studied variance sources in machine learning experiments, finding that data splits often contribute more variance than random seeds or hyperparameters. In high-dimensional settings where $p \approx n$ or $p \gg n$, different training sets may select entirely different variables as important---a problem that motivated stability selection \citep{meinshausen2010stability}. The double descent phenomenon \citep{belkin2019reconciling} illustrates this sensitivity in the interpolating regime: different training sets lead to dramatically different generalization behavior despite identical training performance. Even cross-validation fold assignments affect model selection when performance differences are small relative to fold-to-fold variance \citep{varoquaux2017assessing}.

\paragraph{The Causal Pathway from Noise to Multiplicity.}
Beyond finite-sample effects, noise in outcomes is a fundamental driver of large Rashomon sets. \citet{semenova2023path} establish a causal pathway: (i) outcome noise increases variance of the loss function across data draws; (ii) this increased variance leads to larger generalization gaps; (iii) practitioners respond by constraining to simpler model classes to avoid overfitting; (iv) simpler model classes tend to have larger Rashomon ratios. The implication is that even when the true data-generating process is complex, noise necessitates simpler models for generalization, and these simpler classes exhibit substantial multiplicity. When the signal-to-noise ratio is low, many models achieve similar fit because noise dominates signal \citep{hastie2009elements}; complex models exacerbate this by fitting noise patterns that do not generalize \citep{dietterich1995overfitting}.

\paragraph{Implications for Model Complexity.}
A direct consequence of noise-driven multiplicity is that simple models often match complex ones on real-world problems. \citet{rudin2024amazing} argue that for many tabular data problems, ``there is no accuracy/interpretability trade-off''---simple classification rules, sparse decision trees, and interpretable additive models routinely match random forests and gradient boosted trees. \citet{xin2022exploring} showed that sparse decision trees achieve performance comparable to boosted ensembles on standard benchmarks. \citet{semenova2022existence} found Rashomon ratios exceeding 10\% for stringent tolerance values across benchmark datasets. When such large fractions of models achieve near-optimal performance, the existence of simple, interpretable alternatives within the Rashomon set becomes highly probable.

\paragraph{When Statistical Multiplicity is Small.}
The Rashomon set tends to shrink when labels are generated deterministically from a complex function without noise---approximating such functions exhibits a genuine accuracy-complexity tradeoff \citep{semenova2023path}. Large margins between classes also reduce multiplicity: on well-separated problems like MNIST, most reasonable models succeed because classes are distant in feature space, not because the problem is easy. Statistical multiplicity diminishes when sample size is large relative to model complexity, when signal-to-noise ratio is high, and when the model class contains the true data-generating process. In the limit of infinite data from a well-specified model, this source of multiplicity vanishes---though structural and procedural sources discussed below may persist.

\section{Non-Convexity}
\label{sec:nonconvex}

\textit{This section and the next address structural sources of multiplicity---those arising from the mathematical structure of the learning problem itself. Unlike statistical multiplicity, structural multiplicity persists asymptotically: it cannot be resolved by collecting more data of the same type.}

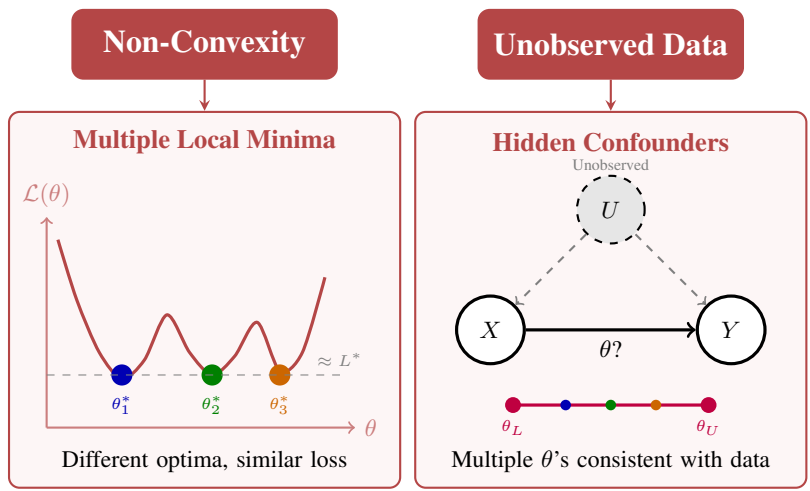
\begin{figure}[htbp]
  \centering
  \begin{adjustbox}{max width=\textwidth}
    \begin{tikzpicture}[
        every node/.style={font=\small},
        arrow/.style={->, thick, >=stealth}
    ]
    
    \definecolor{structred}{RGB}{180, 70, 70}
    \definecolor{structlightred}{RGB}{245, 210, 210}
    
    \node[rectangle, rounded corners, draw=structred, very thick, fill=structred, text=white,
          font=\bfseries\large, minimum width=3.5cm, minimum height=0.9cm] 
        (cause1) at (2.8, 6.2) {Non-Convexity};
    \node[rectangle, rounded corners, draw=structred, very thick, fill=structred, text=white,
          font=\bfseries\large, minimum width=3.5cm, minimum height=0.9cm] 
        (cause2) at (8.2, 6.2) {Unobserved Data};
    
    \node[rectangle, rounded corners, draw=structred, thick, fill=structlightred!20, 
          minimum width=5.2cm, minimum height=5cm] (ill1) at (2.8, 2.8) {};
    \node[font=\bfseries, text=structred] at (2.8, 4.9) {Multiple Local Minima};
    
    \begin{scope}[shift={(0.7, 1.1)}]
        \draw[->, thick, structred!70] (0,0) -- (4.1,0) node[right, font=\footnotesize] {$\theta$};
        \draw[->, thick, structred!70] (0,0) -- (0,2.8) node[above, font=\footnotesize] {$\mathcal{L}(\theta)$};
        
        \draw[structred, very thick, smooth] plot coordinates {
            (0.15, 2.5) (0.4, 1.7) (0.7, 1.0) (1.0, 0.65) (1.3, 0.9) (1.6, 1.5) 
            (1.9, 1.0) (2.2, 0.7) (2.5, 0.95) (2.8, 1.4) (3.1, 0.75) (3.4, 1.0) (3.7, 2.0)
        };
        
        \fill[blue!70!black] (1.0, 0.7) circle (4pt);
        \fill[green!50!black] (2.2, 0.7) circle (4pt);
        \fill[orange!80!black] (3.1, 0.7) circle (4pt);
        
        \node[font=\tiny, blue!70!black] at (1.0, 0.3) {$\theta_1^*$};
        \node[font=\tiny, green!50!black] at (2.2, 0.3) {$\theta_2^*$};
        \node[font=\tiny, orange!80!black] at (3.1, 0.3) {$\theta_3^*$};
        
        \draw[dashed, gray] (0, 0.7) -- (3.9, 0.7);
        \node[font=\tiny, gray] at (3.9, 0.9) {$\approx L^*$};
    \end{scope}
    
    \node[text width=4.5cm, align=center, font=\footnotesize] at (2.8, 0.65) {
        Different optima, similar loss
    };
    
    \node[rectangle, rounded corners, draw=structred, thick, fill=structlightred!20, 
          minimum width=5.2cm, minimum height=5cm] (ill2) at (8.2, 2.8) {};
    \node[font=\bfseries, text=structred] at (8.2, 4.9) {Hidden Confounders};
    
    \begin{scope}[shift={(6.6, 1.6)}]
        \node[circle, draw, very thick, fill=white, minimum size=0.9cm] (X) at (0, 0.8) {$X$};
        \node[circle, draw, very thick, fill=white, minimum size=0.9cm] (Y) at (3.2, 0.8) {$Y$};
        \node[circle, draw, thick, dashed, fill=gray!20, minimum size=0.9cm] (U) at (1.6, 2.4) {$U$};
        
        \draw[->, very thick] (X) -- (Y) node[midway, below, font=\small] {$\theta$?};
        \draw[->, thick, dashed, gray] (U) -- (X);
        \draw[->, thick, dashed, gray] (U) -- (Y);
        
        \node[font=\tiny, gray] at (1.6, 3.0) {Unobserved};
        
        \draw[very thick, purple] (0.3, -0.2) -- (2.9, -0.2);
        \fill[purple] (0.3, -0.2) circle (3pt);
        \fill[purple] (2.9, -0.2) circle (3pt);
        \node[font=\tiny, purple] at (0.3, -0.5) {$\theta_L$};
        \node[font=\tiny, purple] at (2.9, -0.5) {$\theta_U$};
        
        \fill[blue!70!black] (1.0, -0.2) circle (2pt);
        \fill[green!50!black] (1.6, -0.2) circle (2pt);
        \fill[orange!80!black] (2.2, -0.2) circle (2pt);
    \end{scope}
    
    \node[text width=4.5cm, align=center, font=\footnotesize] at (8.2, 0.65) {
        Multiple $\theta$'s consistent with data
    };
    
    \draw[arrow, structred] (cause1.south) -- (ill1.north);
    \draw[arrow, structred] (cause2.south) -- (ill2.north);
    
    \end{tikzpicture}
  \end{adjustbox}
  \caption{\textbf{Structural sources} of the Rashomon effect arise from non-convex optimization landscapes with multiple local minima (left) and unobserved confounders creating non-identifiability (right). This multiplicity persists regardless of sample size.}
  \label{fig:rashomon-structural}
\end{figure}

Convex optimization problems have a fundamental property: any local minimum is also a global minimum, and under mild conditions, this minimum is unique. Most machine learning objectives lack this property. When the loss function $\mathcal{L}(\theta)$ is non-convex in the parameters $\theta$, the optimization landscape admits multiple minima, and different algorithmic trajectories may terminate at different solutions. This multiplicity is not a finite-sample artifact---it reflects the geometry of the problem and persists even with infinite data.

\paragraph{Multiple Minima and Plateaus.}
Non-convex landscapes contain local minima, saddle points, and extended plateaus where the loss varies little across large regions of parameter space \citep{sagun2017empirical}. \citet{choromanska2015loss} analyzed neural network loss surfaces through the lens of statistical physics, finding that the number of local minima grows exponentially with network depth. However, they also found that most local minima have similar loss values, which helps explain an empirical puzzle: neural networks trained from different random initializations typically achieve similar final loss despite converging to detectably different solutions.

\paragraph{Symmetries and Equivalence Classes.}
Some non-convexity arises from symmetries inherent to the model class. Neural networks provide the clearest example: permuting neurons within a hidden layer while correspondingly permuting the weight matrices produces an identical input-output function. A layer with $k$ neurons thus admits $k!$ equivalent parameterizations \citep{sussmann1992uniqueness}. Networks with ReLU activations exhibit additional continuous symmetries---rescaling weights between layers preserves the function \citep{neyshabur2015path}. These symmetries mean that even when the optimal \textit{function} is unique, the optimal \textit{parameterization} is not.

\paragraph{Beyond Prediction}
Non-convex objectives are not confined to neural networks. In causal inference, \citet{parikh2024missing} study the problem of identifying underrepresented subpopulations for which treatment effects cannot be estimated with high precision when generalizing a trial evidence. The resulting variance-minimization objective is non-convex due to ratio structures and binary constraints. Optimizing this objective yields multiple near-optimal solutions that define different subpopulations as ``underrepresented''---each solution implies a different conclusion about who is missing from the study, despite achieving similar statistical precision.

\paragraph{Asymptotic Persistence.}
The key distinction from statistical multiplicity is that non-convexity-induced multiplicity does not vanish with more data. Multiple global minima remain multiple regardless of sample size. \citet{damour2022underspecification} documented this empirically: models indistinguishable by held-out performance exhibited dramatically different behavior under distribution shift, with different predictions on subpopulations not well-represented in training data. They termed this ``underspecification''---training data constrain the model on the observed distribution but leave behavior on other distributions underdetermined. Collecting more data from the same distribution does not resolve this ambiguity \citep{zhang2021understanding}.

\paragraph{Meaningful Diversity in Neural Networks.}
A key challenge in characterizing Rashomon sets for neural networks is that continuous parameter spaces admit infinitely many near-optimal solutions that differ numerically but behave similarly. Simply varying random seeds or applying dropout produces models with different weights but often similar underlying reasoning \citep{damour2022underspecification}. \citet{feng2025rashomon} address this challenge by introducing Rashomon Concept Bottleneck Models (Rashomon CBMs), a framework that learns multiple neural networks which are all accurate yet reason through genuinely distinct human-understandable concepts. Their approach combines lightweight adapter modules with diversity-regularized training to construct diverse models efficiently. Strikingly, their layer-wise analysis reveals that diversity does not emerge uniformly: models share similar representations in early layers but diverge increasingly in deeper layers, suggesting that neural networks can learn low-level features similarly while developing fundamentally different high-level reasoning strategies. On image classification tasks, different models in the Rashomon set identify the same class through entirely different concept combinations---for example, recognizing a tiger by visual appearance (``orange,'' ``stripes''), predatory behavior (``stalker,'' ``hunter''), or environmental context (``jungle,'' ``bush'') \citep{feng2025rashomon}.

\paragraph{The Geometry of Solution Sets.}
Recent work has characterized the geometric structure of non-convex Rashomon sets. Rather than isolated points scattered through parameter space, near-optimal solutions often form connected structures. \citet{garipov2018loss} and \citet{draxler2018essentially} demonstrated that distinct local minima in neural networks are frequently connected by paths of nearly constant loss. \citet{cooper2021global} proved that for sufficiently overparameterized networks, all local minima are global and form a connected manifold. This connectivity means that different training runs sample different points from a continuous set of equivalent solutions---equivalent in training loss, but potentially different in generalization behavior, robustness to perturbation, or interpretability.

\section{Unobserved/Missing Data}
\label{sec:missing}

Missing data, latent variables, and unmeasured confounders create a distinct form of multiplicity: models that are not merely difficult to distinguish statistically but exactly observationally equivalent. This section connects the Rashomon effect to the \textit{partial identification} framework from causal inference and econometrics \citep{manski2003partial, tamer2010partial}, arguing that partial identification represents a limiting case---one where the tolerance $\epsilon$ equals zero and multiplicity persists regardless of sample size.

\paragraph{Partial Identification as Population-Level Rashomon.}
In partial identification, the \textit{identification region} $\Theta_I$ contains all parameter values consistent with the observed data distribution. When $\Theta_I$ is a singleton, the parameter is point identified; otherwise, it is partially identified, and no amount of data can shrink the set further. This identification region is precisely a population-level Rashomon set: every $\theta \in \Theta_I$ achieves identical fit to observable data by construction. Partial identification thus represents the most fundamental form of the Rashomon effect---multiplicity arising not from estimation uncertainty but from what the study design can and cannot reveal about the world.

\paragraph{Unmeasured Confounding.}
Causal inference from observational data provides the canonical example. Let $Y(1)$ and $Y(0)$ denote potential outcomes under treatment and control, with the average treatment effect $\tau = \mathbb{E}[Y(1) - Y(0)]$. The fundamental problem is that we observe $Y = AY(1) + (1-A)Y(0)$, never both potential outcomes for the same unit. When unmeasured variables $U$ influence both treatment $A$ and outcome $Y$, the observed difference $\mathbb{E}[Y \mid A=1] - \mathbb{E}[Y \mid A=0]$ conflates the causal effect with selection bias. \citet{manski1990nonparametric} showed that without restrictions on this confounding, the average treatment effect is bounded to an interval $[\tau_L, \tau_U]$ that additional observational data cannot narrow. Sensitivity analysis \citep{rosenbaum2002observational, cinelli2020making} provides a framework for exploring this Rashomon set: by parameterizing confounding strength, it maps how conclusions vary across the identification region---identifying which conclusions hold across all consistent models and which depend on untestable assumptions about $U$.

\paragraph{Multiplicity in Causal Structure.}
Observational data may also be ambiguous about causal structure itself. \citet{spirtes2000causation} established that causal discovery algorithms identify only Markov equivalence classes---sets of directed acyclic graphs encoding identical conditional independence relations. Three variables satisfying $X \perp Z \mid Y$ are consistent with a chain ($X \to Y \to Z$), a fork ($X \leftarrow Y \to Z$), and other structures implying fundamentally different mechanisms. The effect of intervening on $X$ differs across these structures, yet observations alone cannot distinguish them. \citet{maathuis2009estimating} developed methods to bound causal effects across all DAGs in an equivalence class, yielding intervals reflecting structural ambiguity rather than statistical imprecision.

\paragraph{Missing Data and Latent Variables.}
When data are missing not at random---missingness depends on unobserved values---the complete-data distribution is not identified \citep{little2019statistical}. Multiple distributions are consistent with observed data under different assumptions about the missingness mechanism, and sensitivity analyses map this space \citep{daniels2008missing}. Latent variable models exhibit analogous multiplicities: factor loadings are identified only up to rotation \citep{bollen1989structural}, mixture components only up to label permutation \citep{stephens2000dealing}. These are exact symmetries creating infinite or combinatorially many equivalent parameterizations. In each case, multiple models achieve identical likelihood because the likelihood surface is flat along certain directions in parameter space.

\paragraph{Variable Importance with Unobserved Features.}
Standard variable importance approaches compute importance for a single model using only observed features, but \citet{donnelly2025universe} demonstrate that this can yield misleading conclusions when important variables are omitted. Their framework bounds true variable importance by recognizing that the set of conditional sub-models $S^* = \{f_u : u \in \mathcal{U}\}$---each representing the optimal model for a fixed value of the unobserved variable---is contained within an appropriately expanded Rashomon set. By parameterizing the maximum expected loss $\epsilon_{\text{unobs}}$ of these sub-models, analysts can construct importance bounds that remain valid under specified levels of unobserved confounding, connecting naturally to sensitivity analysis.

\paragraph{The Asymptotic Nature of Identification-Induced Multiplicity.}
The defining feature of this cause is persistence: unlike statistical multiplicity, which shrinks as $n \to \infty$, identification-induced multiplicity reflects fundamental limits of the data type. Resolving it requires either different data---experiments, new measurements, alternative study designs---or additional assumptions that narrow the identification region but cannot be tested against available data. The extensive apparatus developed for partial identification---sharp bounds, sensitivity parameters, impossibility results---offers tools for characterizing Rashomon sets in settings where multiplicity is structural along with statistical.

\section{Limitations of the Optimizer}
\label{sec:optimizer}

\textit{This section and the next address procedural sources of multiplicity---those arising from choices made in the modeling pipeline rather than from properties of the data or problem structure. Procedural multiplicity could in principle be reduced through better algorithms or more computation, though practical constraints often make this infeasible.}

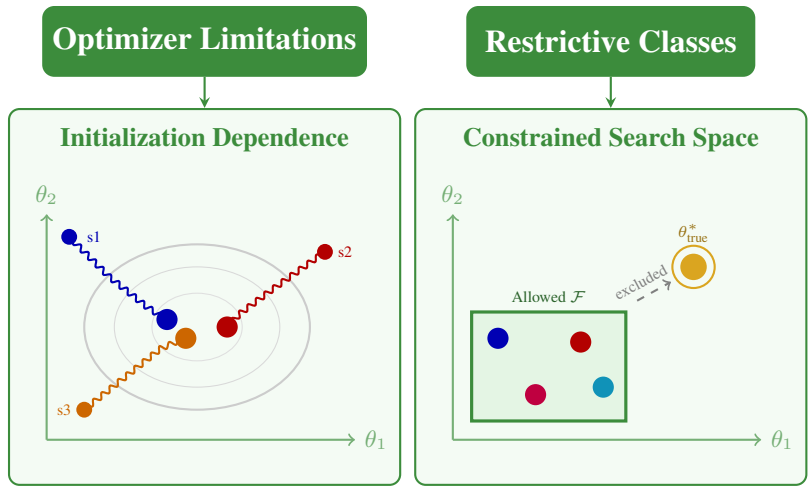
\begin{figure}[htbp]
  \centering
  \begin{adjustbox}{max width=\textwidth}
    \begin{tikzpicture}[
        every node/.style={font=\small},
        arrow/.style={->, thick, >=stealth}
    ]
    
    \definecolor{procgreen}{RGB}{60, 140, 60}
    \definecolor{proclightgreen}{RGB}{200, 235, 200}
    \definecolor{gold}{RGB}{218, 165, 32}
    
    \node[rectangle, rounded corners, draw=procgreen, very thick, fill=procgreen, text=white,
          font=\bfseries\large, minimum width=3.8cm, minimum height=0.9cm] 
        (cause1) at (2.8, 6.2) {Optimizer Limitations};
    \node[rectangle, rounded corners, draw=procgreen, very thick, fill=procgreen, text=white,
          font=\bfseries\large, minimum width=3.8cm, minimum height=0.9cm] 
        (cause2) at (8.2, 6.2) {Restrictive Classes};
    
    \node[rectangle, rounded corners, draw=procgreen, thick, fill=proclightgreen!20, 
          minimum width=5.2cm, minimum height=5cm] (ill1) at (2.8, 2.8) {};
    \node[font=\bfseries, text=procgreen] at (2.8, 4.9) {Initialization Dependence};
    
    \begin{scope}[shift={(0.7, 0.9)}]
        \draw[->, thick, procgreen!70] (0,0) -- (4.1,0) node[right, font=\footnotesize] {$\theta_1$};
        \draw[->, thick, procgreen!70] (0,0) -- (0,3) node[above, font=\footnotesize] {$\theta_2$};
        
        \draw[gray!40, thick] (2.0, 1.5) ellipse (1.5 and 1.1);
        \draw[gray!30] (2.0, 1.5) ellipse (1.1 and 0.8);
        \draw[gray!25] (2.0, 1.5) ellipse (0.6 and 0.45);
        
        \fill[blue!70!black] (0.3, 2.7) circle (3pt);
        \node[font=\tiny, blue!70!black, left] at (0.85, 2.7) {s1};
        \draw[->, blue!70!black, thick, decorate, decoration={snake, amplitude=1pt, segment length=4pt}] 
            (0.3, 2.7) -- (0.8, 2.2) -- (1.3, 1.8) -- (1.6, 1.6);
        \fill[blue!70!black] (1.6, 1.6) circle (4pt);
        
        \fill[red!70!black] (3.7, 2.5) circle (3pt);
        \node[font=\tiny, red!70!black, right] at (3.75, 2.5) {s2};
        \draw[->, red!70!black, thick, decorate, decoration={snake, amplitude=1pt, segment length=4pt}] 
            (3.7, 2.5) -- (3.0, 2.0) -- (2.6, 1.7) -- (2.4, 1.5);
        \fill[red!70!black] (2.4, 1.5) circle (4pt);
        
        \fill[orange!80!black] (0.5, 0.4) circle (3pt);
        \node[font=\tiny, orange!80!black, left] at (0.45, 0.4) {s3};
        \draw[->, orange!80!black, thick, decorate, decoration={snake, amplitude=1pt, segment length=4pt}] 
            (0.5, 0.4) -- (1.0, 0.8) -- (1.5, 1.2) -- (1.85, 1.35);
        \fill[orange!80!black] (1.85, 1.35) circle (4pt);
    \end{scope}
    
    
    \node[rectangle, rounded corners, draw=procgreen, thick, fill=proclightgreen!20, 
          minimum width=5.2cm, minimum height=5cm] (ill2) at (8.2, 2.8) {};
    \node[font=\bfseries, text=procgreen] at (8.2, 4.9) {Constrained Search Space};
    
    \begin{scope}[shift={(6.1, 0.9)}]
        \draw[->, thick, procgreen!70] (0,0) -- (4.1,0) node[right, font=\footnotesize] {$\theta_1$};
        \draw[->, thick, procgreen!70] (0,0) -- (0,3) node[above, font=\footnotesize] {$\theta_2$};
        
        
        \fill[gold] (3.2, 2.3) circle (5pt);
        \draw[gold, thick] (3.2, 2.3) circle (8pt);
        \node[font=\tiny, gold!70!black] at (3.2, 2.75) {$\theta^*_{\text{true}}$};
        
        \draw[procgreen, very thick, fill=proclightgreen!50] 
            (0.25, 0.25) -- (2.3, 0.25) -- (2.3, 1.7) -- (0.25, 1.7) -- cycle;
        \node[font=\tiny, procgreen!70!black] at (1.275, 1.9) {Allowed $\mathcal{F}$};
        
        \fill[blue!70!black] (0.6, 1.35) circle (4pt);
        \fill[red!70!black] (1.7, 1.3) circle (4pt);
        \fill[purple] (1.1, 0.6) circle (4pt);
        \fill[cyan!70!black] (2.0, 0.7) circle (4pt);
        
        \draw[->, dashed, gray, thick] (2.45, 1.85) -- (2.9, 2.1);
        \node[font=\tiny, gray, rotate=28] at (2.5, 2.1) {excluded};
    \end{scope}
    
    
    \draw[arrow, procgreen] (cause1.south) -- (ill1.north);
    \draw[arrow, procgreen] (cause2.south) -- (ill2.north);
    
    \end{tikzpicture}
  \end{adjustbox}
  \caption{\textbf{Procedural sources} of the Rashomon effect arise from optimizer limitations such as initialization dependence (left) and deliberate restrictions to suboptimal model classes (right). These reflect practitioner choices that could in principle be altered.}
  \label{fig:rashomon-procedural}
\end{figure}

Even when a learning problem has a unique optimal solution, practical optimization algorithms may fail to find it. Initialization dependence, stochastic noise, early stopping, and computational constraints all cause different runs to discover different models---each a member of the Rashomon set not because multiple optima exist, but because we cannot reliably locate the one that does.

\paragraph{Initialization and Stochasticity.}
Gradient-based methods depend critically on their starting point. For non-convex objectives, different initializations lead to different basins of attraction and different final solutions \citep{fort2019deep}. Principled initialization schemes \citep{glorot2010understanding, he2015delving} constrain but do not eliminate this sensitivity. Stochastic gradient descent introduces additional variability: mini-batch sampling, data ordering, and gradient noise all cause nominally identical training procedures to yield different models \citep{bengio2012practical}. This stochasticity has regularizing effects---biasing toward flatter minima \citep{keskar2017large}---but the set of flat minima may itself be large, and different noise realizations sample different points within it.

\paragraph{Early Stopping and Checkpoints.}
Practical optimization is never run to convergence. \citet{yao2007early} established that early stopping acts as implicit regularization, with the optimization trajectory tracing a path through model space. Models at different points along this path may all fall within the Rashomon set. \citet{dodge2020fine} showed that for fine-tuning language models, different checkpoint selection criteria yield models with different downstream behavior despite similar aggregate performance.

\paragraph{Approximate Algorithms.}
Many methods do not guarantee finding optima even locally. Greedy decision tree algorithms (CART, ID3, C4.5) select splits optimizing local criteria without regard for global structure \citep{breiman1984classification}. \citet{xin2022exploring} demonstrated that CART rarely produces trees within the Rashomon set of optimal sparse trees---greedy algorithms explore a fundamentally different region of hypothesis space than exact methods. Coordinate descent, EM, and other iterative procedures exhibit analogous path dependence \citep{friedman2010regularization, wu1983convergence}.

\paragraph{Computational Constraints.}
Finite computational budgets force compromises that induce multiplicity. Hyperparameter search explores only a fraction of configuration space; different random seeds find different configurations \citep{bergstra2012random}. Neural architecture search faces the same challenge at larger scale \citep{elsken2019neural}. Distributed training introduces non-determinism through aggregation order and floating-point non-associativity.

\paragraph{Distinguishing Optimizer-Induced from Problem-Induced Multiplicity.}
Optimizer-induced multiplicity differs conceptually from the sources discussed in previous sections: it reflects our inability to find the best model rather than genuine ambiguity about which model is best. In principle, it is eliminable given sufficient computation. In practice, however, optimizer limitations interact with problem structure in complex ways---implicit regularization changes which solutions are preferred, early stopping conflates optimization with model selection, and greedy algorithms systematically explore different regions than exact methods. The models we obtain are shaped jointly by the problem and by our algorithms, making clean attribution of multiplicity to one source or the other difficult without counterfactual reasoning about what perfect optimization would yield.

\section{Restrictive Model Classes}
\label{sec:suboptimal}

Practitioners routinely constrain the hypothesis space---through regularization, architectural choices, or domain requirements---for reasons including overfitting prevention, computational tractability, and interpretability. These restrictions can create multiplicity when the globally optimal model is excluded from consideration: the constrained problem may admit a plateau of near-equivalent approximations rather than a unique minimizer.

\paragraph{Regularization.}
Regularization methods alter the geometry of the solution set in ways that can induce multiplicity. For $\ell_1$-regularized regression (the lasso), \citet{tibshirani1996regression} showed that solutions may not be unique when predictors are correlated, and \citet{zou2005regularization} demonstrated that the lasso arbitrarily selects one variable from correlated groups. Different samples select different variables; small data perturbations switch selections; and the identity of selected variables depends sensitively on the regularization strength $\lambda$. The regularization path $\{\hat{\beta}(\lambda) : \lambda \in [0, \infty)\}$ traces through parameter space, and when cross-validation curves are flat near their minimum, a range of $\lambda$ values---and hence models with different sparsity patterns---are indistinguishable by standard criteria \citep{hastie2009elements}. \citet{meinshausen2010stability} proposed stability selection precisely because the Rashomon set spans multiple points along this path.

\paragraph{Structural Constraints.}
Non-parametric models impose regularization through structural constraints that similarly generate multiplicity. Decision tree complexity is controlled through depth limits, leaf size requirements, and pruning criteria; \citet{rudin2022interpretable} demonstrated that thousands of trees satisfying identical constraints achieve comparable accuracy with different splitting rules and variable usage. The discreteness of tree partitions means many distinct structures approximate the same function when complexity is bounded. Kernel methods exhibit analogous behavior: when cross-validation over bandwidth is flat, a range of bandwidths yield similar performance but different local behavior \citep{wasserman2006all}. Neural network architectures bound capacity through depth, width, and connectivity patterns \citep{hornik1989multilayer, lecun1998gradient, vaswani2017attention}; within each architectural class, multiple parameterizations may achieve similar performance while differing in other properties.

\paragraph{Model Misspecification.}
When the true data-generating process lies outside the assumed model class, multiplicity arises through a different mechanism: multiple models may approximate the truth equally poorly. The ``best'' approximation depends on the covariate distribution, so different samples yield different pseudo-true parameters \citep{white1982maximum, berk1966limiting}. The choice of loss function introduces additional multiplicity under misspecification---models minimizing squared error, absolute error, or Huber loss converge to different limits even asymptotically \citep{huber1967behavior}. This creates methodological multiplicity layered on model multiplicity: the definition of ``best'' itself depends on choices the data cannot adjudicate.

\paragraph{Domain-Driven Constraints.}
Constraints imposed to satisfy domain requirements are particularly likely to induce multiplicity because they are not calibrated to the data. Sparsity constraints requiring at most $k$ features create combinatorial problems where multiple sparse models achieve similar performance, especially when features are correlated or signal is diffuse \citep{bertsimas2016best}. Monotonicity constraints restrict the hypothesis class to functions satisfying shape requirements; when true relationships are only approximately monotonic, multiple constrained models fit similarly well \citep{gupta2016monotonic}. Fairness constraints---demographic parity, equalized odds, calibration---create constrained optimization problems where multiple models achieve similar accuracy-fairness tradeoffs \citep{corbett2017algorithmic}. \citet{coston2021characterizing} showed that the range of achievable fairness levels across accurate models can be substantial.

\paragraph{The Dual Nature of Restrictions.}
Restricting the hypothesis class can either increase or decrease the Rashomon effect depending on the relationship between the restriction and the true data-generating process. Restrictions that exclude the true model tend to create plateaus of similarly-performing approximations. Restrictions that include the true model while excluding poor alternatives can shrink the Rashomon set by eliminating spurious solutions. Since restrictions are typically chosen without knowledge of the truth, the resulting Rashomon set reflects this uncertainty---revealing the extent to which modeling choices, rather than data, determine conclusions.

\section{Discussion}
\label{sec:discussion}

The three categories of causes examined in this paper---statistical, structural, and procedural---are not mutually exclusive. They interact in ways that can amplify or obscure multiplicity. Finite samples blur the distinction between true multiple optima and sampling-induced multiplicity. Noise leads toward simpler model classes that themselves exhibit large Rashomon ratios \citep{semenova2023path}. Non-convex objectives compound the challenges of missing data when algorithms like EM must navigate both non-identifiability and local optima. Regularization choices, driven by finite-sample concerns, determine which region of the hypothesis space is searched. In any given application, the observed Rashomon set reflects the combined influence of all three sources.

\paragraph{Distinguishing the Sources.}
The three-way categorization clarifies what different forms of multiplicity reveal about a learning problem. Statistical multiplicity reflects estimation uncertainty: the Rashomon set shrinks with additional data. Structural multiplicity reflects fundamental limits: the Rashomon set remains large regardless of sample size because non-convexity admits multiple optima or because non-identifiability renders parameters underdetermined by observable data. Procedural multiplicity reflects methodological choices: different optimizers, initializations, or model class specifications yield systematically different solutions. 

\paragraph{Connections Across Literatures.}
A recurring theme is that the Rashomon effect, under various names, has been studied across multiple disciplines. The partial identification literature in econometrics \citep{manski2003partial, tamer2010partial} develops tools for structural multiplicity arising from unobserved data. The sensitivity analysis tradition in biostatistics \citep{rosenbaum2002observational, cinelli2020making} maps how conclusions vary across observationally equivalent models. The underspecification literature in machine learning \citep{damour2022underspecification} documents multiplicity arising from optimization and model class choices. The interpretable machine learning literature \citep{rudin2024amazing, fisher2019all} develops algorithms to explicitly enumerate Rashomon sets. These literatures address the same fundamental phenomenon with complementary tools. Cross-fertilization offers opportunities: partial identification methods can characterize Rashomon sets when multiplicity is structural, while enumeration algorithms can explore high-dimensional identification regions.

\paragraph{Multiplicity and Model Properties.}
When Rashomon sets are large, they often contain models with diverse properties beyond predictive accuracy---varying in complexity, interpretability, fairness, and robustness \citep{rudin2024amazing, semenova2022existence, coston2021characterizing}. This diversity means that the choice among equally-accurate models is consequential: different models in the set may produce different predictions for individuals \citep{marx2020predictive}, suggest different variables as important \citep{fisher2019all}, and behave differently under (unknown) distribution shift \citep{damour2022underspecification}. The existence of many good models also implies that simple, interpretable models often exist within the Rashomon set along with complex black-box alternatives \citep{semenova2022existence, xin2022exploring}.

\paragraph{Future Directions.}
Several questions remain. Efficient characterization of Rashomon sets for high-capacity models like (deep) neural networks is challenging. The behavior of Rashomon sets under distribution shift is not well understood: which models in the set generalize well, and can they be identified before deployment? The connection between partial identification and the Rashomon effect suggests opportunities for methodological exchange, particularly in developing sensitivity analyses that map how conclusions vary across good models. How multiplicity interacts with fairness constraints---and whether fair models cluster in particular regions of the Rashomon set---remains underexplored. Finally, communicating model multiplicity to stakeholders and decision-makers poses challenges that the field has only begun to address.

\paragraph{Summary.}
The Rashomon effect---the existence of many distinct models achieving similar predictive performance---is a fundamental phenomenon in statistical and machine learning modeling. This paper has organized its causes into three categories: statistical sources arising from finite samples and noise, structural sources arising from non-convexity and non-identifiability, and procedural sources arising from optimizer limitations and model class restrictions. These categories differ in whether multiplicity converges asymptotically, persists asymptotically, or reflects engineering decisions. Understanding why many equally good models exist clarifies what conclusions are robust or sensitive to study design choices.

\bibliographystyle{apalike}
\bibliography{references}

@article{breiman2001statistical,
  title={Statistical modeling: The two cultures},
  author={Breiman, Leo},
  journal={Statistical Science},
  volume={16},
  number={3},
  pages={199--231},
  year={2001},
  publisher={Institute of Mathematical Statistics}
}

@article{fisher2019all,
  title={All models are wrong, but many are useful: Learning a variable's importance by studying an entire class of prediction models simultaneously},
  author={Fisher, Aaron and Rudin, Cynthia and Dominici, Francesca},
  journal={Journal of Machine Learning Research},
  volume={20},
  number={177},
  pages={1--81},
  year={2019}
}

@inproceedings{semenova2022existence,
  title={On the existence of simpler machine learning models},
  author={Semenova, Lesia and Rudin, Cynthia and Parr, Ronald},
  booktitle={Proceedings of the 2022 ACM Conference on Fairness, Accountability, and Transparency},
  pages={1827--1858},
  year={2022}
}

@article{rudin2022interpretable,
  title={Interpretable machine learning: Fundamental principles and 10 grand challenges},
  author={Rudin, Cynthia and Chen, Chaofan and Chen, Zhi and Huang, Haiyang and Semenova, Lesia and Zhong, Chudi},
  journal={Statistics Surveys},
  volume={16},
  pages={1--85},
  year={2022}
}

@inproceedings{marx2020predictive,
  title={Predictive multiplicity in classification},
  author={Marx, Charles and Calmon, Flavio and Ustun, Berk},
  booktitle={Proceedings of the 37th International Conference on Machine Learning},
  pages={6765--6774},
  year={2020}
}

@article{damour2022underspecification,
  title={Underspecification presents challenges for credibility in modern machine learning},
  author={D'Amour, Alexander and Heller, Katherine and Moldovan, Dan and Adlam, Ben and Alber, Brendan and Baldassarre, Federico and Banber, Bibi and others},
  journal={Journal of Machine Learning Research},
  volume={23},
  number={226},
  pages={1--61},
  year={2022}
}

@inproceedings{watson2023predictive,
  title={Predictive and causal implications of using {S}hapley value for model interpretation},
  author={Watson, David S and Gultchin, Limor and Taly, Ankur and Floridi, Luciano},
  booktitle={Proceedings of the 2023 AAAI/ACM Conference on AI, Ethics, and Society},
  pages={228--238},
  year={2023}
}

@inproceedings{black2022model,
  title={Model multiplicity: Opportunities, concerns, and solutions},
  author={Black, Emily and Yeom, Samuel and Fredrikson, Matt},
  booktitle={Proceedings of the 2022 ACM Conference on Fairness, Accountability, and Transparency},
  pages={850--863},
  year={2022}
}

@article{rudin2019stop,
  title={Stop explaining black box machine learning models for high stakes decisions and use interpretable models instead},
  author={Rudin, Cynthia},
  journal={Nature Machine Intelligence},
  volume={1},
  number={5},
  pages={206--215},
  year={2019}
}

@inproceedings{coston2021characterizing,
  title={Characterizing fairness over the set of good models under selective labels},
  author={Coston, Amanda and Mishler, Alan and Kennedy, Edward H and Chouldechova, Alexandra},
  booktitle={Proceedings of the 38th International Conference on Machine Learning},
  pages={2144--2155},
  year={2021}
}

@article{ioannidis2005most,
  title={Why most published research findings are false},
  author={Ioannidis, John PA},
  journal={PLoS Medicine},
  volume={2},
  number={8},
  pages={e124},
  year={2005}
}

@article{fort2019deep,
  title={Deep ensembles: A loss landscape perspective},
  author={Fort, Stanislav and Hu, Huiyi and Lakshminarayanan, Balaji},
  journal={arXiv preprint arXiv:1912.02757},
  year={2019}
}

@book{manski2003partial,
  title={Partial Identification of Probability Distributions},
  author={Manski, Charles F},
  year={2003},
  publisher={Springer}
}

@article{tamer2010partial,
  title={Partial identification in econometrics},
  author={Tamer, Elie},
  journal={Annual Review of Economics},
  volume={2},
  number={1},
  pages={167--195},
  year={2010}
}

@article{bouthillier2021accounting,
  title={Accounting for variance in machine learning benchmarks},
  author={Bouthillier, Xavier and Delaunay, Pierre and Bronzi, Mirko and Trofimov, Assya and Nichyporuk, Brennan and Szeto, Justin and others},
  journal={Proceedings of Machine Learning and Systems},
  volume={3},
  pages={747--769},
  year={2021}
}

@article{meinshausen2010stability,
  title={Stability selection},
  author={Meinshausen, Nicolai and B{\"u}hlmann, Peter},
  journal={Journal of the Royal Statistical Society: Series B (Statistical Methodology)},
  volume={72},
  number={4},
  pages={417--473},
  year={2010}
}

@article{belkin2019reconciling,
  title={Reconciling modern machine-learning practice and the classical bias--variance trade-off},
  author={Belkin, Mikhail and Hsu, Daniel and Ma, Siyuan and Mandal, Soumik},
  journal={Proceedings of the National Academy of Sciences},
  volume={116},
  number={32},
  pages={15849--15854},
  year={2019}
}

@article{varoquaux2017assessing,
  title={Assessing and tuning brain decoders: Cross-validation, caveats, and guidelines},
  author={Varoquaux, Ga{\"e}l and Raamana, Pradeep Reddy and Engemann, Denis A and Hoyos-Idrobo, Andr{\'e}s and Schwartz, Yannick and Thirion, Bertrand},
  journal={NeuroImage},
  volume={145},
  pages={166--179},
  year={2017}
}

@book{hastie2009elements,
  title={The Elements of Statistical Learning: Data Mining, Inference, and Prediction},
  author={Hastie, Trevor and Tibshirani, Robert and Friedman, Jerome},
  year={2009},
  publisher={Springer},
  edition={2nd}
}

@inproceedings{choromanska2015loss,
  title={The loss surfaces of multilayer networks},
  author={Choromanska, Anna and Henaff, Mikael and Mathieu, Michael and Arous, G{\'e}rard Ben and LeCun, Yann},
  booktitle={Proceedings of the 18th International Conference on Artificial Intelligence and Statistics},
  pages={192--204},
  year={2015}
}

@article{sagun2017empirical,
  title={Empirical analysis of the {H}essian of over-parameterized neural networks},
  author={Sagun, Levent and Evci, Utku and Guney, V Ugur and Dauphin, Yann and Bottou, Leon},
  journal={arXiv preprint arXiv:1706.04454},
  year={2017}
}

@article{sussmann1992uniqueness,
  title={Uniqueness of the weights for minimal feedforward nets with a given input-output map},
  author={Sussmann, Hector J},
  journal={Neural Networks},
  volume={5},
  number={4},
  pages={589--593},
  year={1992}
}

@inproceedings{neyshabur2015path,
  title={Path-{SGD}: Path-normalized optimization in deep neural networks},
  author={Neyshabur, Behnam and Salakhutdinov, Ruslan R and Srebro, Nathan},
  booktitle={Advances in Neural Information Processing Systems},
  volume={28},
  year={2015}
}

@article{zhang2021understanding,
  title={Understanding deep learning (still) requires rethinking generalization},
  author={Zhang, Chiyuan and Bengio, Samy and Hardt, Moritz and Recht, Benjamin and Vinyals, Oriol},
  journal={Communications of the ACM},
  volume={64},
  number={3},
  pages={107--115},
  year={2021}
}

@inproceedings{garipov2018loss,
  title={Loss surfaces, mode connectivity, and fast ensembling of {DNN}s},
  author={Garipov, Timur and Izmailov, Pavel and Podoprikhin, Dmitrii and Vetrov, Dmitry P and Wilson, Andrew G},
  booktitle={Advances in Neural Information Processing Systems},
  volume={31},
  year={2018}
}

@inproceedings{draxler2018essentially,
  title={Essentially no barriers in neural network energy landscape},
  author={Draxler, Felix and Veschgini, Kambis and Salmhofer, Manfred and Hamprecht, Fred},
  booktitle={Proceedings of the 35th International Conference on Machine Learning},
  pages={1309--1318},
  year={2018}
}

@article{cooper2021global,
  title={Global minima of overparameterized neural networks},
  author={Cooper, Yaim},
  journal={SIAM Journal on Mathematics of Data Science},
  volume={3},
  number={2},
  pages={676--691},
  year={2021}
}

@article{manski1990nonparametric,
  title={Nonparametric bounds on treatment effects},
  author={Manski, Charles F},
  journal={American Economic Review},
  volume={80},
  number={2},
  pages={319--323},
  year={1990}
}

@book{rosenbaum2002observational,
  title={Observational Studies},
  author={Rosenbaum, Paul R},
  year={2002},
  publisher={Springer},
  edition={2nd}
}

@article{cinelli2020making,
  title={Making sense of sensitivity: Extending omitted variable bias},
  author={Cinelli, Carlos and Hazlett, Chad},
  journal={Journal of the Royal Statistical Society: Series B (Statistical Methodology)},
  volume={82},
  number={1},
  pages={39--67},
  year={2020}
}

@book{spirtes2000causation,
  title={Causation, Prediction, and Search},
  author={Spirtes, Peter and Glymour, Clark N and Scheines, Richard and Heckerman, David},
  year={2000},
  publisher={MIT Press},
  edition={2nd}
}

@article{maathuis2009estimating,
  title={Estimating high-dimensional intervention effects from observational data},
  author={Maathuis, Marloes H and Kalisch, Markus and B{\"u}hlmann, Peter},
  journal={Annals of Statistics},
  volume={37},
  number={6A},
  pages={3133--3164},
  year={2009}
}

@book{bollen1989structural,
  title={Structural Equations with Latent Variables},
  author={Bollen, Kenneth A},
  year={1989},
  publisher={John Wiley \& Sons}
}

@article{stephens2000dealing,
  title={Dealing with label switching in mixture models},
  author={Stephens, Matthew},
  journal={Journal of the Royal Statistical Society: Series B (Statistical Methodology)},
  volume={62},
  number={4},
  pages={795--809},
  year={2000}
}

@book{little2019statistical,
  title={Statistical Analysis with Missing Data},
  author={Little, Roderick JA and Rubin, Donald B},
  year={2019},
  publisher={John Wiley \& Sons},
  edition={3rd}
}

@book{daniels2008missing,
  title={Missing Data in Longitudinal Studies: Strategies for {B}ayesian Modeling and Sensitivity Analysis},
  author={Daniels, Michael J and Hogan, Joseph W},
  year={2008},
  publisher={CRC Press}
}

@inproceedings{glorot2010understanding,
  title={Understanding the difficulty of training deep feedforward neural networks},
  author={Glorot, Xavier and Bengio, Yoshua},
  booktitle={Proceedings of the 13th International Conference on Artificial Intelligence and Statistics},
  pages={249--256},
  year={2010}
}

@inproceedings{he2015delving,
  title={Delving deep into rectifiers: Surpassing human-level performance on {ImageNet} classification},
  author={He, Kaiming and Zhang, Xiangyu and Ren, Shaoqing and Sun, Jian},
  booktitle={Proceedings of the IEEE International Conference on Computer Vision},
  pages={1026--1034},
  year={2015}
}

@incollection{bengio2012practical,
  title={Practical recommendations for gradient-based training of deep architectures},
  author={Bengio, Yoshua},
  booktitle={Neural Networks: Tricks of the Trade},
  pages={437--478},
  year={2012},
  publisher={Springer}
}

@inproceedings{keskar2017large,
  title={On large-batch training for deep learning: Generalization gap and sharp minima},
  author={Keskar, Nitish Shirish and Mudigere, Dheevatsa and Nocedal, Jorge and Smelyanskiy, Mikhail and Tang, Ping Tak Peter},
  booktitle={Proceedings of the International Conference on Learning Representations},
  year={2017}
}

@article{yao2007early,
  title={On early stopping in gradient descent learning},
  author={Yao, Yuan and Rosasco, Lorenzo and Caponnetto, Andrea},
  journal={Constructive Approximation},
  volume={26},
  number={2},
  pages={289--315},
  year={2007}
}

@article{dodge2020fine,
  title={Fine-tuning pretrained language models: Weight initializations, data orders, and early stopping},
  author={Dodge, Jesse and Ilharco, Gabriel and Schwartz, Roy and Farhadi, Ali and Hajishirzi, Hannaneh and Smith, Noah},
  journal={arXiv preprint arXiv:2002.06305},
  year={2020}
}

@book{breiman1984classification,
  title={Classification and Regression Trees},
  author={Breiman, Leo and Friedman, Jerome and Stone, Charles J and Olshen, Richard A},
  year={1984},
  publisher={CRC Press}
}

@article{friedman2010regularization,
  title={Regularization paths for generalized linear models via coordinate descent},
  author={Friedman, Jerome and Hastie, Trevor and Tibshirani, Rob},
  journal={Journal of Statistical Software},
  volume={33},
  number={1},
  pages={1--22},
  year={2010}
}

@article{wu1983convergence,
  title={On the convergence properties of the {EM} algorithm},
  author={Wu, C F Jeff},
  journal={Annals of Statistics},
  volume={11},
  number={1},
  pages={95--103},
  year={1983}
}

@article{bergstra2012random,
  title={Random search for hyper-parameter optimization},
  author={Bergstra, James and Bengio, Yoshua},
  journal={Journal of Machine Learning Research},
  volume={13},
  number={2},
  pages={281--305},
  year={2012}
}

@article{elsken2019neural,
  title={Neural architecture search: A survey},
  author={Elsken, Thomas and Metzen, Jan Hendrik and Hutter, Frank},
  journal={Journal of Machine Learning Research},
  volume={20},
  number={55},
  pages={1--21},
  year={2019}
}

@article{tibshirani1996regression,
  title={Regression shrinkage and selection via the lasso},
  author={Tibshirani, Robert},
  journal={Journal of the Royal Statistical Society: Series B (Methodological)},
  volume={58},
  number={1},
  pages={267--288},
  year={1996}
}

@article{zou2005regularization,
  title={Regularization and variable selection via the elastic net},
  author={Zou, Hui and Hastie, Trevor},
  journal={Journal of the Royal Statistical Society: Series B (Statistical Methodology)},
  volume={67},
  number={2},
  pages={301--320},
  year={2005}
}

@inproceedings{xin2022exploring,
  title={Exploring the whole {R}ashomon set of sparse decision trees},
  author={Xin, Rui and Zhong, Chudi and Chen, Zhi and Takagi, Takuya and Seltzer, Margo and Rudin, Cynthia},
  booktitle={Advances in Neural Information Processing Systems},
  volume={35},
  pages={14071--14084},
  year={2022}
}

@book{wasserman2006all,
  title={All of Nonparametric Statistics},
  author={Wasserman, Larry},
  year={2006},
  publisher={Springer}
}

@article{hornik1989multilayer,
  title={Multilayer feedforward networks are universal approximators},
  author={Hornik, Kurt and Stinchcombe, Maxwell and White, Halbert},
  journal={Neural Networks},
  volume={2},
  number={5},
  pages={359--366},
  year={1989}
}

@article{lecun1998gradient,
  title={Gradient-based learning applied to document recognition},
  author={LeCun, Yann and Bottou, L{\'e}on and Bengio, Yoshua and Haffner, Patrick},
  journal={Proceedings of the IEEE},
  volume={86},
  number={11},
  pages={2278--2324},
  year={1998}
}

@inproceedings{vaswani2017attention,
  title={Attention is all you need},
  author={Vaswani, Ashish and Shazeer, Noam and Parmar, Niki and Uszkoreit, Jakob and Jones, Llion and Gomez, Aidan N and Kaiser, {\L}ukasz and Polosukhin, Illia},
  booktitle={Advances in Neural Information Processing Systems},
  volume={30},
  year={2017}
}

@article{white1982maximum,
  title={Maximum likelihood estimation of misspecified models},
  author={White, Halbert},
  journal={Econometrica},
  volume={50},
  number={1},
  pages={1--25},
  year={1982}
}

@article{berk1966limiting,
  title={Limiting behavior of posterior distributions when the model is incorrect},
  author={Berk, Robert H},
  journal={Annals of Mathematical Statistics},
  volume={37},
  number={1},
  pages={51--58},
  year={1966}
}

@inproceedings{huber1967behavior,
  title={The behavior of maximum likelihood estimates under nonstandard conditions},
  author={Huber, Peter J},
  booktitle={Proceedings of the Fifth Berkeley Symposium on Mathematical Statistics and Probability},
  volume={1},
  number={1},
  pages={221--233},
  year={1967}
}

@article{bertsimas2016best,
  title={Best subset selection via a modern optimization lens},
  author={Bertsimas, Dimitris and King, Angela and Mazumder, Rahul},
  journal={Annals of Statistics},
  volume={44},
  number={2},
  pages={813--852},
  year={2016}
}

@article{gupta2016monotonic,
  title={Monotonic calibrated interpolated look-up tables},
  author={Gupta, Maya and Cotter, Andrew and Pfeifer, Jan and Voevodski, Konstantin and Canini, Kevin and Mangylov, Alexander and Mber, Neil and Narber, Sanja},
  journal={Journal of Machine Learning Research},
  volume={17},
  number={109},
  pages={1--47},
  year={2016}
}

@inproceedings{corbett2017algorithmic,
  title={Algorithmic decision making and the cost of fairness},
  author={Corbett-Davies, Sam and Pierson, Emma and Feller, Avi and Goel, Sharad and Huq, Aziz},
  booktitle={Proceedings of the 23rd ACM SIGKDD International Conference on Knowledge Discovery and Data Mining},
  pages={797--806},
  year={2017}
}

@article{dietterich1995overfitting,
  title={Overfitting and undercomputing in machine learning},
  author={Dietterich, Tom},
  journal={ACM Computing Surveys},
  volume={27},
  number={3},
  pages={326--327},
  year={1995}
}

@inproceedings{semenova2023path,
  title={A path to simpler models starts with noise},
  author={Semenova, Lesia and Chen, Harry and Parr, Ronald and Rudin, Cynthia},
  booktitle={Advances in Neural Information Processing Systems},
  volume={36},
  year={2023}
}

@inproceedings{rudin2024amazing,
  title={Amazing things come from having many good models},
  author={Rudin, Cynthia and Zhong, Chudi and Semenova, Lesia and Seltzer, Margo and Parr, Ronald and Liu, Jiachang and Katta, Srikar and Donnelly, Jon and Chen, Harry and Boner, Zachery},
  booktitle={Proceedings of the 41st International Conference on Machine Learning},
  year={2024},
  organization={PMLR}
}

@article{coker2021theory,
  title={A theory of statistical inference for ensuring the robustness of scientific results},
  author={Coker, Beau and Rudin, Cynthia and King, Gary},
  journal={Management Science},
  volume={67},
  number={10},
  pages={6174--6197},
  year={2021}
}

@inproceedings{cooper2024arbitrariness,
  title={Arbitrariness and prediction: The confounding role of variance in fair classification},
  author={Cooper, A Feder and Lee, Katherine and Choksi, Madiha and Barocas, Solon and De, Christopher and Grimmelmann, James and Kleinberg, Jon and Sen, Siddharth and Zhang, Baobao},
  booktitle={Proceedings of the AAAI Conference on Artificial Intelligence},
  volume={38},
  pages={22004--22012},
  year={2024}
}

@article{hsu2022rashomon,
  title={Rashomon capacity: A metric for predictive multiplicity in classification},
  author={Hsu, Hsiang and Calmon, Flavio},
  booktitle={Advances in Neural Information Processing Systems},
  volume={35},
  pages={28988--29000},
  year={2022}
}

@inproceedings{donnelly2023rashomon,
  title={The {R}ashomon importance distribution: Getting {RID} of unstable, single model-based variable importance},
  author={Donnelly, Jon and Katta, Srikar and Rudin, Cynthia and Browne, Edward P},
  booktitle={Advances in Neural Information Processing Systems},
  year={2023}
}

@inproceedings{zhong2023exploring,
  title={Exploring and interacting with the set of good sparse generalized additive models},
  author={Zhong, Chudi and Chen, Zhi and Liu, Jiachang and Seltzer, Margo and Rudin, Cynthia},
  booktitle={Advances in Neural Information Processing Systems},
  year={2023}
}

@inproceedings{liu2022fasterrisk,
  title={{F}aster{R}isk: Fast and accurate interpretable risk scores},
  author={Liu, Jiachang and Zhong, Chudi and Li, Boxuan and Seltzer, Margo and Rudin, Cynthia},
  booktitle={Advances in Neural Information Processing Systems},
  year={2022}
}

@inproceedings{wang2022timbertrek,
  title={{T}imber{T}rek: Exploring and curating sparse decision trees with interactive visualization},
  author={Wang, Zijie J and Zhong, Chudi and Xin, Rui and Takagi, Takuya and Chen, Zhi and Chau, Duen Horng and Rudin, Cynthia and Seltzer, Margo},
  booktitle={IEEE Visualization and Visual Analytics (VIS)},
  pages={60--64},
  year={2022}
}

@article{wang2021gamchanger,
  title={Gam changer: Editing generalized additive models with interactive visualization},
  author={Wang, Zijie J and Kale, Alex and Nori, Harsha and Stella, Peter and Nunnally, Mark and Chau, Duen Horng and Vorvoreanu, Mihaela and Vaughan, Jennifer Wortman and Caruana, Rich},
  journal={arXiv preprint arXiv:2112.03245},
  year={2021}
}

@article{parikh2024missing,
author = {Harsh Parikh and Rachael K. Ross and Elizabeth Stuart and Kara E. Rudolph},
title = {Who Are We Missing?: A Principled Approach to Characterizing the Underrepresented Population},
journal = {Journal of the American Statistical Association},
volume = {120},
number = {551},
pages = {1414--1423},
year = {2025},
publisher = {Taylor \& Francis},
doi = {10.1080/01621459.2025.2495319},
}

@article{donnelly2025universe,
  title={Doctor Rashomon and the UNIVERSE of Madness: Variable Importance with Unobserved Confounding and the Rashomon Effect},
  author={Donnelly, Jon and Katta, Srikar and Borgonovo, Emanuele and Rudin, Cynthia},
  journal={arXiv preprint arXiv:2510.12734},
  year={2025}
}

@inproceedings{feng2025rashomon,
  title={Many Ways to be Right: {R}ashomon Sets for Concept-Based Neural Networks},
  author={Feng, Shihan and Zhang, Cheng and Xi, Michael and Hsu, Ethan and Semenova, Lesia and Zhong, Chudi},
  booktitle={arXiv preprint arXiv:2511.19636},
  year={2025}
}

\appendix
\section{Algorithms for Exploring Rashomon Sets}
\label{sec:algorithms}

A major recent development is the creation of algorithms that explicitly enumerate or efficiently represent entire Rashomon sets for certain model classes. These algorithms represent a paradigm shift from seeking a single optimal model to characterizing the full space of near-optimal solutions.

\subsection{TreeFARMS: Rashomon Sets of Decision Trees}

\citet{xin2022exploring} developed TreeFARMS (Tree Finding Algorithm for Rashomon Model Sets), which finds all sparse decision trees within a specified loss tolerance. Key features include:

\begin{itemize}
    \item Exact enumeration of all trees satisfying accuracy and sparsity constraints
    \item Efficient representation using lattice structures
    \item Ability to handle datasets with thousands of observations and tens of features
    \item Support for multiple objectives (accuracy, F1-score, fairness metrics)
\end{itemize}

For the COMPAS recidivism dataset, TreeFARMS found 1,365 sparse decision trees achieving near-optimal accuracy, each with different splitting rules and variable usage patterns.

\subsection{GAM Rashomon Sets}

\citet{zhong2023exploring} developed algorithms for finding Rashomon sets of sparse generalized additive models (GAMs). Their approach:

\begin{itemize}
    \item Computes the number of unique support sets (feature subsets used)
    \item Represents the Rashomon set through convex coefficient regions for each support
    \item Enables interactive exploration through the GAMChanger tool \citep{wang2021gamchanger}
\end{itemize}

\subsection{FasterRisk: Rashomon Sets of Scoring Systems}

\citet{liu2022fasterrisk} developed FasterRisk for finding accurate sparse scoring systems---integer-weighted linear models commonly used in medicine and criminal justice. The algorithm efficiently searches over:

\begin{itemize}
    \item Different feature subsets
    \item Different integer coefficient values
    \item Different intercepts and scaling factors
\end{itemize}

\subsection{ROOT: Rashomon Sets for Causal Inference}

\citet{parikh2024missing} introduced ROOT (Rashomon Set of Optimal Trees), a functional optimization approach. It has been operationalized for characterizing underrepresented populations in trial generalization. ROOT:

\begin{itemize}
    \item Uses tree-sampling with explore-exploit strategy
    \item Produces interpretable characterizations of populations where treatment effects can be precisely estimated
    \item Generates a Rashomon set of near-optimal trees for population characterization
\end{itemize}

\subsection{Interactive Exploration Tools}

Complementing these algorithms, interactive visualization tools enable practitioners to explore Rashomon sets:

\begin{itemize}
    \item \textbf{TimberTrek} \citep{wang2022timbertrek}: Visualizes and enables navigation of decision tree Rashomon sets, allowing users to filter by accuracy, features used, and other criteria.
    
    \item \textbf{GAMChanger} \citep{wang2021gamchanger}: Enables domain experts to directly manipulate GAM weights while maintaining accuracy, effectively exploring the Rashomon set interactively.
\end{itemize}

\end{document}